\documentclass[11pt]{article}
\usepackage{enumitem}
\usepackage[round]{natbib}
\usepackage{authblk}
\usepackage{graphicx} 
\usepackage{amsmath}
\usepackage{amsfonts}
\usepackage{hyperref}
\usepackage[export]{adjustbox}
\usepackage{fancyhdr}
\usepackage{lastpage}
\usepackage{relsize} 
\usepackage{microtype}
\usepackage{xcolor}
\usepackage{comment}

\newcommand{\dia}{\mathlarger{\mathlarger{\mathlarger{\diamond}}}}
\newcommand{\red}[1]{\textcolor{black}{#1}}
\title{Taking Principles Seriously: A Hybrid Approach \\to Value Alignment}

\author[1]{Tae Wan Kim}
\author[1]{John Hooker}
\author[2]{Thomas Donaldson}
\affil[1]{Carnegie Mellon University, USA}
\affil[2]{University of Pennsylvania, USA}

\begin{document}

\maketitle

\begin{abstract}
\noindent An important step in the development of value alignment (VA) systems in AI is understanding how VA can reflect valid ethical principles.  We propose that designers of VA systems incorporate ethics by utilizing a \mbox{hybrid} approach in which both ethical reasoning and empirical observation play a role.  This, we argue, avoids committing ``naturalistic fallacy,'' which is an attempt to derive ``ought'' from ``is,'' and it provides a more adequate form of ethical reasoning when the fallacy is not committed.  Using quantified model logic, we precisely formulate principles derived from deontological ethics and show how they imply particular ``test propositions'' for any given action plan in an AI rule base.  The action plan is ethical only if the test proposition is empirically true, a judgment that is made on the basis of empirical VA.  This permits empirical VA to integrate seamlessly with independently justified ethical principles.
\end{abstract}

\section{Introduction}
Artificial intelligence (AI) technologies increasingly replace human decision makers. Worries rise about the compatibility of AI and human values. A growing number of researchers are examining how AI can acquire moral intelligence \citep{wallach2008moral, burton2017ethical, walsh2019effective, allen2011robot, bringsjord2013robots, scheutz2016we, arnold2017value, arnold2018big}. Such attempts are recently lumped under the term ``value alignment'' (hereafter VA).   \citeauthor{russell2015research}\ (\citeyear{russell2015research}) highlight the need for VA and identify two options for achieving it:

\begin{quote}
    ``[A]ligning the values of powerful AI systems with our own values and preferences ... [could involve either] a system [that] infers the preferences of another rational or nearly rational actor by observing its behavior ... [or] could be explicitly inspired by the way humans acquire ethical values.'' 
\end{quote} 

\noindent As this passage suggests, one option for VA is to teach machines human preferences, and another is to teach machines ethics. The word ``values'' in fact has this double meaning.  It can refer to what humans value in the sense of what they see as subjectively preferable, or it can refer to reasonably defensible ethical principles. The distinction is important, because we acquire knowledge of the two types of values in different ways.  

A similar distinction occurs in previous literature under the names {\em top-down} and \mbox{\em bottom-up} VA \citep{allen2000prolegomena,allen2002calculated,AllenSmitWallach2005,allen2006machine,wallach2008moral,wallach2008machine}.  \citeauthor{russell2015research}\ suggest a bottom-up approach in the form of inverse reinforcement learning, which allows a machine to \red{internalize} a pattern of preferences by observing how humans actually behave \citep{abbeel2004apprenticeship,ng2000algorithms}.  Reinforcement learning, and machine learning (ML) in general, \red{offer a number of advantages but must deal with such issues as inadequate reward functions to represent complex ethical norms, biased data, and opaqueness} \citep{arnold2017value, prince1988rules, marcus2018deep}. A promising alternative to ML is {\em logic-based} VA, \red{which} has received less attention despite having a long research record \citep{arkoudas2005toward,bringsjord2006toward,bringsjord2012divine,bringsjord201721st, govindarajulu2017automating, kim2018toward}.

In this paper, we make a case for {\em hybrid} VA that combines  ML-based and logic-based approaches. 
A logic-based approach is especially important because it allows the use of  ``independently justified'' or ``independently defensible'' ethical principles.  By these we mean principles that find their justification in ethical theory.  Such principles are ``normative'' in the sense commonly used by moral philosophers: they are prescriptive rather than descriptive and are elements of traditional normative moral theories such as deontology, consequentialism and virtue ethics.  Such principles are increasingly discussed as candidates for  computational use \citep{lindner2018formalization, lindner2020evaluation, ganascia2007modelling}. Independently justified principles avoid many problems, including those associated with the well-known {\em is-ought gap}, one aspect of which is reflected in the unconscious biases now widely studied by behavioral ethicists \citep{bazerman2011blind}. In turn, we propose our own version of deontological VA for use in such a hybrid approach. 

\red{After elaborating on why a purely ML-based approach is inadequate, we show how symbolic logic enables the introduction of deontological reasoning into machine ethics.  Rather than opting for a particular version of moral theory, we attempt to develop a comprehensive, ecumenical framework of ethical principles \citep{parfit2011matters}. We first articulate univeralization, utilitarian, and autonomy-based principles  in the idiom of quantified modal logic.  We then use these principles to derive {\em test propositions}, also formulated in modal logic, for each action specified by an AI rule base.  The action is ethical only if the test propositions are empirically true, a judgment that can be based on machine learning and empirical VA.  This permits empirical VA to integrate seamlessly with independently justified ethical principles.  }

\section{Two different value alignment systems}
AI is an imitation game. It imitates the human mind. Because more than one theory of mind is possible, different models of AI are are also possible, and so too,  different models of VA.  Broadly speaking, two categories of VA  stand out, ML-based and  logic-based, although neither is instantiated perfectly in any given working AI system (Table~\ref{tab:my-table}).

\begin{table}[!t]
\centering
\caption{\red{Comparison of ML-based and logic-based VA}}
\label{tab:my-table}
\vspace{1.5ex} 
\resizebox{\textwidth}{!}{%
\begin{tabular}{|l|l|l|}
\hline
 & & \\[-1.5ex]
  & \textbf{ML-based} & \textbf{Logic-based}  \\[0.7ex] 
\hline\hline
 & & \\[-2ex]
\textbf{Theory of mind} & Connectionism & Computationalism \\[0.7ex] \hline
 & & \\[-2ex]
\textbf{Base discipline} & Statistics & Logic \\[0.7ex] \hline
 & & \\[-2ex]
\textbf{AI techniques}                                                  & \begin{tabular}[c]{@{}l@{}}Machine learning \\ (automated statistics, deep learning) \end{tabular}
& \begin{tabular}[c]{@{}l@{}}Symbolic AI \\ (i.e., GOFAI: Good Old Fashioned AI)\end{tabular}                                                            \\[2.3ex] \hline
 & & \\[-2ex]
\textbf{Value alignment}   & Bottom-up & Top-down \\[0.7ex]  \hline
 & & \\[-2ex]
\textbf{Example}                                                        & \begin{tabular}[c]{@{}l@{}}ML system trained by lay people's \\ perception of fairness regarding \\ autonomous vehicles and \\ gender/racial discrimination.\end{tabular} & \begin{tabular}[c]{@{}l@{}}Formalized normative principles \\ (e.g., double effect theory, categorical \\ imperatives) using symbolic logic \\ (e.g., quantified modal logic)\end{tabular} \\[5.3ex] \hline
 & & \\[-2ex]
\textbf{\begin{tabular}[c]{@{}l@{}}Dual process \\ theory\end{tabular}} 
& System 1 & System 2 \\[2.3ex] \hline
\end{tabular}%
}
\end{table}

ML-based VA is connectionist. Connectionism holds that human intelligence can be explained and imitated by using artificial neural nets consisting of three kinds of connected units: input, hidden and output \citep{sep-connectionism}. Deep Learning (DL) exemplifies connectionism by utilizing a complex ``automated statistics'' based on a large number of hidden and opaque heuristics using associations \citep{Danks}. ML's major advantage, which is especially obvious in an end-to-end model such as DL, is its powerful ability to imitate and further strengthen skill sets in training data. DL has illustrated the power of connectionist models by capably learning human skills, especially in the domain of pattern recognition, face recognition, medical diagnostic systems, and text reading.

\subsection{The is-ought gap and the problem of bounded ethicality}

Since connectionist systems are inductive, the quality of ML-based VA  relies heavily on that of inputs. If training data is biased or unethical, the system will generate well-imitated, undesirable outputs. Microsoft's AI-based chatter-bot Tay (an acronym for ``thinking about you'') was designed to engage with people on Twitter and learn from them how to carry on a conversation. When some people started tweeting racist and misogynistic expressions, Tay responded in kind. Microsoft immediately terminated the experiment \citep{Wolf:2017:WWS:3144592.3144598}. Most algorithmic bias problems we see now are results of ML-based VA, which uses data sets from humans who already have implicit or explicit biases. 

These mistakes reflect an error well-known to moral philosophers, the problem of deriving an ``ought'' from an ``is,''  sometimes called the ``naturalistic fallacy.''  From the fact that people behave in racist ways,  it cannot follow that people ought to behave in such ways. While not a formal fallacy, the violation of the is-ought gap signals a form of epistemic na\"{i}vet\'{e}, one that ignores the axiom in normative ethics that ``no justifiable `ought' can be derived directly from an `is'.''  Disagreements about the robustness of the fallacy abound \citep{donaldson1994integration,donaldson2012epistemic,pigden2016hume,woods2017model}, and so this paper adopts a modest, workable interpretation of the is-ought gap coined recently by Daniel Singer, namely, ``There are no valid arguments from non-normative premises to a relevantly normative conclusion'' \citep{singer2015mind}. Descriptive (or naturalistic) statements are reportive of what states of affairs are like, whereas normative statements are stipulative and action-guiding. Examples of the former are ``The grass is green'' and ``Many people find deception to be unethical.'' Examples of the latter are ``You ought not murder'' and ``Lying is unethical.'' Normative statements usually figure in the semantics of deontic (obligation-based) or evaluative expressions such as ``ought,'' ``impermissible,'' ``wrong,'' ``good,'' ``bad,'' or ``unethical.''  One may object that a high-level domain-general premise such as ``machines ought to have our values no matter what'' might successfully link a descriptive premise to a normative conclusion.  This objection is correct, but allows the original problem to pop up again at a deeper level.  ``What facts,'' one might ask, ``justify the conclusion that machines ought to imitate perfectly our behaviors?'' 

Using data from ``unbiased'' people's behaviors seems an obvious solution, but the problem is more complicated than one thinks. The preceding decade of research in behavioral ethics has shown the existence of various pernicious influences on ethical decisions, often at an unconscious level. When these influences lead to unethical behavior that conflict with an actor's moral beliefs and commitments \citep{moore2006conflicts}, the phenomenon is often referred to as ``bounded ethicality''\citep{bazerman2011bounded, bazerman2011blind, chugh2005bounded, tenbrunsel2005bounded}. One example of bounded ethicality is ``ordinary prejudice,'' which reveals itself in implicit associations about gender, race, and other demographic groups \citep{bertrand2005implicit, green2007implicit, greenwald2009understanding, rudman2007discrimination}. These associations can lead to unintentionally discriminatory results, such as discriminatory hiring practices and unwarranted discrepancies in the evaluation of the skills and competencies of workers. Other elements of bounded ethicality include ``in-group favoritism,'' ``self-serving bias,'' and ``motivated blindness,'' the last of which refers to a systemic but unconscious failure to notice unethical behavior in oneself or others even when it is in one's financial interest to do so \citep{bazerman2011time, moore2010conflict}. One might consider using professional moral philosophers' opinions as training data for ML-based VA \citep{AndAnd11},  but recent research shows both expert judgment generally and ethical expert judgment in particular to be frequently biased.  Professional ethicists' moral intuitions and specific  judgements turn out to be as vulnerable to biases or irrelevant factors as those of lay persons \citep{schwitzgebel2012expertise, wiegmann2020intuitive, tobia2013moral, schwitzgebel2015philosophers, egler2020philosophical}.  Because any attempt to use the ML-based VA system to generate the principles would be viciously circular, ML-based systems stand in need of independently defensible principles in order to evaluate even the training data to be used.   

Logic-based VA is distinct from the ML-based in several ways. It is analogous to computationalism, in which human intelligence operates as a computer does, or in other words, in step with a set of systematic, abstract, symbol-and-rule mechanisms that are transparently expressed with formal-symbolic logic \citep{sep-computational-mind, scheutz2002computationalism}. Due to the popularity of ML systems, logic-based systems are \red{sometimes referred to as} GOFAI (``good old-fashioned AI'') \citep{haugeland1985artificial}. But logic-based AI is still widely used, for instance, in the driving mechanisms of autonomous drones or cars, even though the pattern recognition mechanisms in these applications are primarily based on ML systems.  Logic-based approaches are especially useful when formalizing independently defensible ethical principles of the sort invoked by professional philosophers.   Such logic-based systems are  sometimes labeled {\em symbolic AI}. Interestingly enough, formal logic is one of a few languages shared by both computer scientists and moral philosophers.  Unlike eliminative (pure) connectionist systems, logic-based VA relies not on associations, but on deductive logic and logical proofs.

\subsection{The problem of System 2 and systematicity}

From a psychological perspective, ML systems are relevantly similar to what dual process theory \citep{kahneman2011thinking} knows as ``System~1'' \citep{chauvet201830, geffner2018model, rossi2019preferences}.  It is opaque, fast, and intuitive to use. Dual process theory frames the human mind in terms of two distinctive processes: System~1 and System~2.  In contrast to System~1 thinking, System~2 thinking is slow, transparent, analytical, logical, reasons-responsive and computational.  Research shows that unethical and biased decisions are correlated with System~1 thinking, and that shifting the mode to System~2 thinking is often an effective way to avoid unethical behaviors \citep{bazerman2012behavioral, bazerman2016bounded, zhang2015reducing, sezer2015ethical}.  This is despite the fact that System~1 thinking can be useful in other domains where intuitive associations are useful, such as in making heuristic decisions. 

Because ML systems draw upon System~1 behavior, ML-based VA can be inherently vulnerable to unethical decision making. The ``systematicity'' challenge, neglected by connectionists for decades \citep{calvo2014architecture, lake2018still, alhama2019review, geffner2018model, marcus2001algebraic}, sheds further light on this.  In 1988, linguistic philosophers \citep{fodor1988connectionism} argued that connectionism confuses the intrinsically systematic nature of thought with a system of associations. More specifically, they argued that thoughts---e.g., ``Mary loves John''---must involve operations with a set of rules (e.g., syntactic and semantic combinatorial relations or grammars). Pure or eliminative connectionist systems, which rely exclusively on associations,  lack the ability to employ rules, and this seriously limits their ability to explain human thinking.  A human who can think  ``Mary loves John'' can also think ``John loves Mary,'' but purely connectionist systems trained by connectionist methods cannot systematically do the latter without further resources. Responding to this challenge, many connectionists have attempted to show that structured ML systems might be redesigned, but the attempts underscore the eventual need for pure or eliminative ML systems that employ rule-like structures. 

Our purpose here is not to adjudicate this debate. However, the debate itself reveals the need for connectionist systems to be used within their legitimate scope. In that sense, our view is roughly consistent with that of Paul Smolensky who responded to the systematicity challenge in his article, ``On the proper treatment of connectionism (PTC)'' (\citeyear{smolensky1988proper}). Similar to dual process theory, Smolensky's ``proper treatment of connectionism'' construes human intelligence in terms of two distinct realms: on the one hand, there is ``cultural knowledge'' (e.g., formalized knowledge presented by symbols and rule-like logic), and on the other, there is ``individual knowledge'' (e.g., perception, intuitive processing).   Connectionist systems are adequate for the latter, but not the former. The proper treatment of connectionism entails that computational systems are necessary but insufficient for language-like processing because human language operates against a backdrop of empirical, common-sensical knowledge which, in turn, allows rules themselves to make sense.  

This broad point is especially relevant for moral thought, in which the ``reasoning'' portion of moral thinking relies upon systemic operations instead of associations.  A person who can reason, ``It is wrong for Jane to gratuitously lie to Mary'' can also reason ``It is wrong for Mary to gratuitously lie to Jane'' or ``It is not wrong for \ldots.''  Moral reasoning is fundamentally rule-based. It can be said that a person who concludes ``It is wrong for Jane to lie to Mary'' uses a rule such as ``It is wrong for agent x to gratuitously  lie to someone'' and an empirical premise, ``Jane gratuitously lies to Mary.''\footnote{Moral particularism criticizes rule-based ethical theory, but grants easily that rules are  used in moral reasoning, even as it critiques a one-size-fits-all approach.  Interestingly, a rule-based or logic-based ethical theory is not committed to to a rigorous one-size-fits-all approach \citep{smith2011understanding}.}

\section{\red{Related work}}

Bringsjord and his collaborators \citep{bringsjord2006toward,bringsjord2012divine, bringsjord201721st, arkoudas2005toward,govindarajulu2017automating} are the first we know to use deontic logic to explicitly represent philosophically justifiable ethical principles such as the doctrine of double effect. Our approach dovetails with that of Bringsjord in identifying the importance of deontic logic for teaching right and wrong to machines. Since his pioneering work, many others have attempted to represent  ethical principles using deontic logic. These contributions reveal the versatility of deontic logic when formalizing not only deontological moral theory but other traditions, such as areteic theory (including virtue ethics) and commandment theory.

Our work is consistent with the established deontic tradition in moral philosophy that uses deontic logic to formalize deontological moral theory.  Rather than opting for a particular version of moral theory, we attempt to develop a comprehensive, ecumenical framework of ethical principles\citep{parfit2011matters}. We offer a deontological representation of three central ethical traditions, using a generalization principle, an autonomy principle, and a deontic utility principle. Using deontology, we \red{indicate in outline} how ethical obligations can be derived from first principles instead of relying on conflicting \mbox{moral intuitions} of what seems fair or unbiased. While ethical philosophy has been viewed as vague and subjective by the popular imagination, the deontological approach to moral philosophy is known for offering a  rigorous foundation.  

\citeauthor{AllenSmitWallach2005}  first suggested a hybrid approach to VA and recommended combining top-down and bottom-up approaches.  Although their distinction can be more broadly construed, a typical top-down approach installs ethical principles directly into the machine, while a bottom-up approach typically asks the machine to learn prescriptive norms from experience.  From an epistemological perspective, the typical bottom-up VA approach can result in teaching strategies that sometimes conflate ``is'' and ``ought.''  For example, one might suggest that a machine might learn ethics through a simulated process of evolution \citep{conitzer2017moral}.  The fact that certain ethical norms evolve does not imply that they are valid ethical principles \citep{berker2009normative, nagel1979ethics, mcdowell1995two, rachels1990created}. \red{It is true that bottom-up approach does not automatically commit the naturalistic fallacy, particularly if ethical principles validate the norms learned in this fashion \citep{wallach2008moral}.}
\red{Nonetheless, in our approach to hybrid VA, bottom-up learning does none of the normative work, but is used only to evaluate the truth of test propositions derived from ethical principles. }

Another version of a hybrid approach to VA is advocated by \citeauthor[p. 81]{arnold2017value} who argue, ``architectures must explicitly represent legal, ethical and moral principles,'' while using them as principles for decision-making in order to achieve predictable decisions on the part of the system. Systems that uphold those principles as much as possible represent a more ethical path than systems that are less transparent less accountably trained, and less easily corrected.’’ We largely agree with these authors, and our efforts are indebted to their insightful criticism of the IRL-based VA. \citeauthor{arnold2017value} suggest that the problems in the IRL approach can be significantly addressed by an hybrid approach in which explicitly written ethical rules can be imposed as constraints on what a machine learns from observation through IRL. We follow this very path by developing deontological principles as constraints, realizing nonetheless that one must ask precisely what remains within the unconstrained space of observational learning.  If what remains is learning that includes ethical norms, then  once again we confront the is-ought gap.  If, on the other hand, it is learning that includes empirical facts about the world, then those facts alone cannot be transformed into ``oughts.''


\red{It is with this in mind that} we offer a hybrid approach to VA that integrates independently justified ethical principles from the deontological tradition in ethics \citep{Korsgaard1996, Nagel1986, ONeill2014} with factual knowledge \red{acquired through ML technology.}   Relevant facts may include observed preferences and values, but even such value-relevant facts cannot be the source of ethical principles.


Applying the imperative, ``Thou shalt not kill,'' to a given action requires at a minimum that someone knows the facts relevant to the action \citep{hare1991language}.  The relevant facts, which may include observations of human values and preferences, \red{do not by themselves decide what is ethical, but they factor into ethical assessment.  
In addition, action decisions almost always take the form, ``If the facts are such-and-such, then do A,'' which we refer to as an {\em action plan}.  }  
This provides a clue as to how VA can knit together empirical observation and ethical principles.  \red{The factual information in an action plan can be merged with ethical imperatives that depend on factual circumstances to arrive at an ethical judgment.  The next section describes in detail how this can be accomplished. }

\section{Integrating ethical principles and empirical VA}

We now show how deontologically derived ethical principles can combine with empirical facts in a systematic way.  An adequate exposition of deontological reasoning is far beyond the scope of this paper, and we do not attempt to defend the specific ethical principles we have chosen, although we briefly explain why we think they are reasonable.  Relevant literature is cited for readers who wish to study the underlying arguments in detail.  Our purpose here is only to show how a careful statement of ethical principles clarifies how these principles can interrelate with empirical observation in VA.  

We argue that expressing ethical assertions in the idiom of quantified modal logic, as developed in \cite{kim2018toward}, makes the relationship between ethical principles and empirical observation perspicuous.  \red{Specifically: {\em ethical principles imply certain logical propositions that must be true in order for a given action plan to be ethical, and empirical observation determines whether these propositions are, in fact, true.}  We refer these propositions as {\em test propositions}, whose empirical evaluation typically requires  observation of human values, beliefs, and behavior.  The test propositions need not appear alongside the action plans in an AI system, but they can be generated and evaluated automatically if desired (Section~4.5).}

Thus the role of ethics in hybrid VA is to derive necessary conditions for the rightness of specific actions, and the role of empirical VA is to ascertain whether these conditions are satisfied in the real world.

\subsection{Actions and Reasons} 

Deontology derives ethical principles from the logical structure of action \citep{Kan1785,wood1999kant,ONeill2014,HooKim19a}. It begins with the necessity of distinguishing free action from mere behavior, insofar as causally speaking, both are determined by chemical and physical forces.  
Contemporary deontological thinkers usually base the distinction between free and causally determined behavior on a Kantian {\em dual standpoint} theory of ethics that identifies free action as behavior for which the agent has {\em reasons} \citep{Bil06,Korsgaard1996,Nagel1986,Nel00}.  Such reasons are not themselves psychological causes or motivations, but considerations that the agent consciously makes to justify a choice.  The reasons need not be good or convincing ones from another agent's perspective, but must be sufficiently coherent to serve as an explanation of why the agent chose the action.   

Ethical principles are necessary conditions for the coherence or intelligibility of the reasons behind an action.  While a number of necessary conditions for coherence are possible, ethical principles rest on the {\em universality of reason}: an agent who takes a set of reasons as justifying an action must in order to be consistent take the reasons as justifying the same action for any agent to whom those reasons apply.  

We  focus on the three ethical principles that have been most intensely studied in the literature---generalization, utility maximization, and respect for autonomy.  Each states a necessary condition for ethical conduct.  We make no claim that they are exhaustive, but only that they illustrate how empirical VA can be anchored by ethical principles.   

Before proceeding, two caveats are in order.  First, in this paper we do not attempt to convince readers of the superiority of the deontological tradition or its premise that principles can be discovered through an analysis of the logical structure of action.  Our aim is more modest: to show that deontology is particularly suitable for hybrid VA. Two of the three principles we employ,  generalization and respect for autonomy, \red{have historical roots in} Kant's The Formula of the Universal Law and The Formula of Humanity \citep{wood1999kant}, although our formulations of them \red{differ}. Second, we also use a deontic model of utilitarianism (e.g., \citeauthor{cummiskey1996kantian}, \citeyear{cummiskey1996kantian}) in order to make utilitarianism consistent with the other two other principles.

\subsection{Generalization Principle}

The universality of reason leads immediately to the {\em generalization principle}:  a rational agent must believe that his/her reasons for acting are consistent with the assumption that all rational agents to whom the reasons apply could engage in the same  actions \citep{ONeill2014,wood1999kant}.  

As an example, suppose I see wristwatches on open display in a shop and steal one.  My reasons for the theft are that I would like to have a new watch, and that I can get away with taking one.\footnote{In practice, the reasons for theft are likely to be more complicated than this.  I may be willing to steal partly because I believe the shop can easily withstand the loss, no employee will be disciplined or terminated due to the loss, I will not feel guilty afterward, and so forth.  But for purposes of illustration we suppose there are only two reasons.}  At the same time, I cannot rationally believe that I would be able to get away with the theft if {\em everyone} stole watches when these reasons apply.  The shop would install security measures to prevent theft, which is inconsistent with one of my reasons for stealing the watch.  The theft therefore violates the generalization principle.

To give these ideas more precision, we express the action plan and generalization principle in the language of quantified modal logic.  In so doing, we do not define a deductive system or propose formal semantics, as they are unnecessary for our project.  We merely borrow logical notation in order to allow a more rigorous formulation and application of ethical principles.  

The decision to steal a watch can be expressed in logical notation as follows.  Define predicates
\[
\begin{array}{l}
C_1(a) = \mbox{Agent $a$ would like to possess an item on} \\
\hspace{8.5ex} \mbox{display in a shop.} \\
C_2(a) = \mbox{Agent $a$ can get away with stealing the item.} \\
A_1(a) = \mbox{Agent $a$ will steal the item.}
\end{array}
\]
Because the agent's reasons are an essential part of moral assessment, we evaluate the agent's {\em action plan}, which states that the agent will take a certain action when certain reasons apply.  In this case, the action plan is
\begin{equation}
\big(C_1(a)\wedge C_2(a)\big) \Rightarrow_a A_1(a)  \label{eq:action}
\end{equation}
Here $\Rightarrow_a$ is not logical entailment but indicates that agent $a$ regards $C_1(a)$ and $C_2(a)$ as justifying $A_1(a)$.  The reasons in the action plan should be the most general set of conditions that the agent takes as justifying the action.  Thus the action plan refers to an item in a shop rather than specifically to a watch, because the fact that it is a watch is not relevant to the justification; what matters is whether the agent wants the item and can get away with stealing it.

We can now state the generalization principle using quantified modal logic.  Let $C(a)\Rightarrow_a A(a)$ be an action plan for agent $a$, where $C(a)$ is a conjunction of the reasons for taking action $A(a)$.  The action plan is generalizable if and only if
\begin{equation}
\begin{array}{l}
\dia_a P\Big( \forall x \big(C(x)\Rightarrow_x A(x)\big) \wedge C(a)\wedge A(a)  \Big)
\end{array} \label{eq:gen0}
\end{equation}
Here $P(S)$ means that it is possible for proposition $S$ to be true, and $\dia_a S$ means that $a$ can rationally believe $S$.  The proposition $\dia_a S$ is equivalent to $\neg \Box_a \neg S$, where $\Box_a \neg S$ means that rationality requires require $a$ to deny $S$.\footnote{The operators $\dia$ and $\Box$ have a somewhat different interpretation here than in traditional epistemic and doxastic modal logics, but the identity $\dia S \equiv \neg\Box\neg S$ holds as usual.}  Thus (\ref{eq:gen0}) says that agent $a$ can rationally believe that it is possible for everyone to have the same action plan as $a$, even while $a$'s reasons still apply and $a$ takes the action.  

Returning to the theft example, the condition (\ref{eq:gen0}) \red{becomes the test proposition for action plan (\ref{eq:gen0})}:
\begin{equation}
\begin{array}{l}
\dia_a P\Big( \forall x \big(C_1(x)\wedge C_2(x)\Rightarrow_x A_1(x)\big) \wedge C_1(a)\wedge C_2(a)\wedge A_1(a)  \Big)
\end{array} \label{eq:gen}
\end{equation}
This says that it is rational for $a$ to believe that it is possible for the following to be true simultaneously: (a) everyone steals when the stated conditions apply, and (b) the conditions apply and $a$ steals.
Since (\ref{eq:gen}) is false, action plan (\ref{eq:action}) is unethical.

The necessity of (\ref{eq:gen}) for the rightness of action plan (\ref{eq:action}) is anchored in deontological theory, while the falsehood of (\ref{eq:gen}) is a fact about the world.  This fact might be inferred by collecting responses from shop owners about how they  would react if theft were widespread.  {\em Thus ethics and empirical VA work together in a very specific way: ethics tells us that \red{the test proposition} (\ref{eq:gen}) must be true if the theft is to be ethical, and  empirical VA provides evidence that bears on whether (\ref{eq:gen}) is true.}

An action plan in the autonomous vehicle domain might be 
\begin{equation}
C_3(a) \Rightarrow_a A_2(a) \label{eq:gen4}
\end{equation}
where 
\[
\begin{array}{l}
C_3(a) = \mbox{An ambulance under the control of agent $a$ can reach its}\\
\hspace{11ex} \mbox{destination sooner by using siren and lights.} \\
A_2(a) = \mbox{Agent $a$ will direct an ambulance to use siren and lights.}
\end{array}
\]
Agent $a$ is the ambulance driver, or in the case of an autonomous vehicle, the designer of the software that controls the ambulance.  The generalization principle \red{yields the test proposition}
\begin{equation}
{\dia}_a P\Big(\forall x \big( C_3(x)\Rightarrow_y A_2(x) \big) \wedge C_3(a) \wedge A_2(a) \Big)
\label{eq:gen5}
\end{equation}
This says that it is rational for agent $a$ to believe that siren and lights could continue to hasten arrival if all ambulances used them for all trips, emergencies and otherwise.  If empirical VA reveals that most drivers would ignore siren and lights if they were universally abused in this fashion, then we have evidence that (\ref{eq:gen5}) is false, in which case action plan (\ref{eq:gen4}) is unethical.

\subsection{Maximizing Utility}

Utilitarianism is normally understood as a {\em consequentialist} theory that evaluates an act by its actual consequences.  Specifically, an act is ethical only if it maximizes total net expected utility across all who are affected.  Yet the utilitarian principle can also be construed in a deontological fashion \citep{cummiskey1996kantian}, which allows it to be interpreted as requiring the agent to select actions that the agent can rationally believe will maximize utility.  
While utilitarians frequently view utility maximization as the sole ethical principle, it can be seen as an additional necessary condition for an ethical action.  The other non-utilitarian principles remain in force because only actions that satisfy the other principles are considered options for maximizing utility.

In a deontological analysis, utility is not what people generally value but what the agent is rationally committed to valuing.
The logic of means and ends requires that the agent regard some end as {\em intrinsically} valuable (such as happiness), and the universality of reason requires that it be seen as valuable for any agent.  A utilitarian believes this commits the agent to selecting actions that maximize the expected net sum of utility over everyone who is affected.\footnote{Alternatively, one might argue that maximizing the minimum utility over those affected (or achieving a lexicographic maximum) is the rational way to take everyone's utility into account, after the fashion of John Rawls's difference principle \citep{rawls2009theory}.  Or one might argue for some rational combination of utilitarian and equity objectives \citep{KarMor15,hooker2012combining}.  However, for many practical applications, simple utility maximization appears to be a sufficiently close approximation to a ``rational'' choice, and to simplify exposition we assume so in this paper. }


The utilitarian principle can be formalized \red{by requiring that a given action plan create at least as much utility as any other available action plan}.  Let $u(C(a),A(a))$ be a utility function that measures the total net expected utility of action $A(a)$  under conditions $C(a)$.  Then an action plan $C(a)\Rightarrow_a A(a)$ satisfies the utilitarian principle only if agent $a$ can rationally believe that action $A(a)$ creates at least as much utility as any ethical action that is  available under the same circumstances.  This can be written
\begin{equation}
{\dia}_a \forall A' \Big( E\big(C(a),A'(a)\big) \rightarrow u\big(C(a),A(a)\big) \geq u\big( C(a),A'(a)\big) \Big) \label{eq:util0}
\end{equation}
where $A'$ ranges over actions.  The predicate $E(C(a),A'(a))$ means that action $A'(a)$ is available for agent $a$ under conditions $C(a)$, and that the action plan $C(a)\Rightarrow_a A'(a)$ is generalizable and respects autonomy.\footnote{For ``respecting autonomy,'' see the next section.}  Note that we are now quantifying over predicates and have therefore moved into second-order logic.

Popular views about acceptable behavior frequently play a role in applications of the utilitarian principle.  For example, in some parts of the world, drivers consider it wrong to enter a stream of moving traffic from a side street without waiting for a gap in the traffic.  In other parts of the world this can be acceptable, because drivers in the main thoroughfare expect it and make allowances.   Suppose driver $a$'s action plan is $(C_4(a)\wedge C_5(a))\Rightarrow_a A_3(a)$, where
\[
\begin{array}{l}
C_4(a) = \mbox{Driver $a$ wishes to enter a main thoroughfare.} \\
C_5(a) = \mbox{Driver $a$ can enter a main thoroughfare by moving} \\
\hspace{9ex} \mbox{into the traffic without waiting for a gap.}\\
A_3(a) = \mbox{Driver $a$ will move into traffic without waiting} \\
\hspace{9ex} \mbox{for a gap.}
\end{array}
\]
As before, driver $a$ is the designer of the software if the vehicle is autonomous. Using (\ref{eq:util0}), the driver's action plan maximizes utility only if \red{the following test proposition is true}:
\begin{equation}
\begin{array}{l}
{\displaystyle
{\dia}_a \forall A' \Big( E\big(C_4(a),C_5(a),A'(a)\big) \rightarrow 
} \\ 
\hspace{15ex} 
{\displaystyle
u\big(C_4(a),C_5(a),A_3(a)\big) \geq u\big( C_4(a),C_5(a),A'(a)\big) \Big)
}
\end{array} \label{eq:util1}
\end{equation}
Suppose we wish to design driving policy in a context where pulling immediately into traffic is considered unacceptable.  Then doing so is a dangerous move that no one is expecting, and an accident could result.  Waiting for a gap in the traffic results in greater net expected utility, or formally, $u(C_4(a),C_5(a),A_3(a))<u(C_4(a),C_5(a),A_4(a))$, where $A_4(a)$ is the action of moving into traffic after waiting for a gap.  So (\ref{eq:util1}) is false, and its falsehood can be inferred by collecting popular views about acceptable driving behavior.  Observed preferences and values are therefore relevant to an ethical assessment, but they alone do not determine the assessment.

{\em Again we have a clear demonstration of how ethical principles can combine with empirical VA.  The utilitarian principle tells us that a particular action plan is ethical only if \red{test proposition} (\ref{eq:util1}) is true, and empirical VA tells us whether (\ref{eq:util1}) is true.}

\red{A similar approach can accommodate other situations in which popular expectations bear on ethical decisions.  For example, it has been observed that people may expect different ethical norms to be followed by machine agents than by humans \citep{MalSchArnVoiCos15}.  This could affect generalizability as well as a utilitarian assessment, because there may be different implied promises or agreements concerning machines than humans.  Yet again, expectations alone do not determine the ethical outcome.}

\subsection{Respect for Autonomy}

A third ethical principle requires agents to respect the autonomy of other agents.  Specifically, an agent should not adopt an action plan that the agent is rationally constrained to believe is inconsistent with an ethical action plan of another agent, without informed consent.  Murder, enslavement, and inflicting serious injury are extreme examples of autonomy violations because they interfere with many ethical action plans.  Coercion may or may not violate autonomy, depending on precisely how action plans are formulated.\footnote{A more adequate analysis leads to a principle of {\em joint} autonomy, according to which it is violation of autonomy to adopt an action plan that is mutually inconsistent with action plans of a set of other agents, when those other action plans are themselves mutually consistent.  Joint autonomy addresses situations in which an action necessarily interferes with the action plan of some agent but no particular agent, as when someone throws a bomb into a crowd.  A general formulation of the joint autonomy principle in terms of modal operators is given in \cite{kim2018toward}.  This and other complications are discussed in \cite{Hooker2018}. }

The argument for respecting autonomy is basically as follows.  Suppose I violate someone's autonomy for certain reasons.  That person could, at least conceivably, have the same reasons to violate my autonomy.  This means that, due to the universality of reason, I am endorsing the violation of my own autonomy in such a case.  This is a logical contradiction, because it implies that I am deciding not to do what I decide to do.  To avoid contradicting myself, I must avoid interfering with other action plans.

To formulate an autonomy principle, we say that agent $a$'s action plan $C(a)\Rightarrow_a A(a)$ is consistent with $b$'s action plan $C'(b)\Rightarrow_b A'(b)$ when 
\begin{equation}
{\dia}_a P \big( A(a) \wedge A'(b)\big) \; \vee \;
\neg {\Box}_a P\big( C(a)\wedge C'(b)\big) 
\label{eq:ac0}
\end{equation}
This says that agent $a$ can rationally believe that the two actions are mutually consistent, or can rationally believe that the reasons for the actions are mutually inconsistent.  The latter suffices to avoid inconsistency of the action plans, because if the reasons for them cannot both apply, the actions can never come into conflict.  

As an example of how coercion need not violate autonomy, suppose agent $b$ wishes to catch a bus and has decided to cross the street to a bus stop, provided no traffic is coming.  The agent's action plan is
\begin{equation}
\big( C_6(b) \wedge C_7(b) \wedge \neg C_8(b)\big) \Rightarrow_b A_5(b)
\label{eq:ac1}
\end{equation}
where 
\[
\begin{array}{l}
C_6(b) = \mbox{Agent $b$ wishes to catch a bus.}\\
C_7(b) = \mbox{There is a bus stop across the street from $b$.} \\
C_8(b) = \mbox{There are cars approaching $b$.} \\
A_5(b) = \mbox{Agent $b$ will cross the street.} 
\end{array}
\]
Agent $a$ sees agent $b$ begin to cross the street and forcibly pulls $b$ out of the path of an oncoming car that $b$ does not notice.  Agent $a$'s action plan is
\begin{equation}
\big( C_8(b) \wedge C_9(b)\big) \Rightarrow_a A_6(a,b) 
\label{eq:ac2}
\end{equation}
where
\[
\begin{array}{l}
C_9(b) = \mbox{Agent $b$ is about to cross the street.}\\
A_6(a,b) = \mbox{Agent $a$ will prevent agent $b$ from crossing the street.} 
\end{array}
\]
Agent $a$ does not violate agent $b$'s autonomy, even though there is coercion.  Their action plans (\ref{eq:ac1}) and (\ref{eq:ac2}) are consistent with each other, because the condition (\ref{eq:ac0}) \red{yields the test proposition}
\begin{equation}
{\dia}_a P \big( A_5(b) \wedge A_6(a,b)\big) \; \vee \;
\neg {\Box}_a P\big( C_6(b) \wedge C_7(b) \wedge \neg C_8(b) \wedge C_8(b) \wedge C_9(b)\big) 
\label{eq:ac3}
\end{equation}
This means that either (a) agent $a$ can rationally believe that the two actions are consistent with each other, or (b) agent $a$ can rationally believe that the antecedents of (\ref{eq:ac1}) and (\ref{eq:ac2}) are mutually inconsistent.  As it happens, the two actions are obviously not consistent with each other, and so (a) is false.  However, agent $a$ can rationally believe that the antecedents of (\ref{eq:ac1}) and (\ref{eq:ac2}) are mutually inconsistent, because $C_8(b)$ and $\neg C_8(b)$ are contradictory.  This means (b) is true, which implies that condition (\ref{eq:ac3}) is satisfied, and there is no violation of autonomy.  

{\em Again, this clearly distinguishes the roles of ethics and empirical observation in VA.  Ethical reasoning tells us that \red{the test proposition} (\ref{eq:ac3}) must be true if autonomy is to be respected, whereas observation of the world tells us whether (\ref{eq:ac3}) is true.}

\red{In saying that coercion can be ethical, we do not imply that a violation of autonomy can be ethical.  Coercion must be consistent with the coerced agent's action plan, as in the above example.  Coercion can also be ethical when there is implied or informed consent, or when it is necessary to prevent unethical behavior, as in self-defense.\footnote{Coercion can be ethical when there is informed consent to a risk of interference, because giving informed consent is equivalent to including the possibility of interference as one of the antecedents of the action plan. \red{This occurs, for example, when a medical test subject gives consent with the knowledge that an experimental drug may cause illness, even though administering a drug that turns out to be harmful is a form of coercion.}  Interfering with an unethical action plan is no violation of autonomy because an unethical action plan is, strictly speaking, not an action plan due to the absence of a coherent set of reasons for undertaking it.  An action plan is considered unethical in this context when it violates the generalization or utility principle, or interferes with an action plan that does not violate one of these principles, and so on recursively.  \red{Thus coercion is ethical in an act of self-defense, or to stop someone from unethically harming others.}} }

To illustrate how autonomy may play a role in the ethics of driving, suppose that a pedestrian $b$ dashes in front of $a$'s rapidly moving car.  Driver $a$ can slam on the brake and avoid impact with the pedestrian, but another driver $c$ is following closely, and a sudden stop could cause a crash.  
The driver $a$ must choose between two possible action plans:
\begin{align}
& \big(C_{10}(a,b) \wedge C_{11}(a,c)\big) \Rightarrow_a A_7(a) \label{eq:ac10} \\
& \big(C_{10}(a,b) \wedge C_{11}(a,c)\big) \Rightarrow_a \neg A_7(a) \label{eq:ac11}
\end{align}
where
\[
\begin{array}{l}
C_{10}(a,b) = \mbox{Pedestrian $b$ is dashing in front of $a$'s car.}\\
C_{11}(a,c) = \mbox{Driver $c$ is closely following $a$'s car.} \\
A_7(a) = \mbox{Agent $a$ will immediately slam on the brake.} 
\end{array}
\]
Meanwhile, the pedestrian $b$ has any number of action plans that are clearly inconsistent with death or serious injury.  Let $C_{12}(b) \Rightarrow_b A_8(b)$ be one of them.
Also driver $c$ of the other car (there is only one occupant) has action plans that are inconsistent with an injury.  We suppose that $C_{13}(c)\Rightarrow_c A_9(c)$ is one of them.

We first check whether hitting the brake, as in action plan (\ref{eq:ac10}), is inconsistent with the other driver's action plan $C_{13}(c)\Rightarrow_c A_9(c)$.  \red{The test proposition is}
\begin{equation}
{\dia}_a P \big( A_7(a) \wedge A_9(c) \big) \; \vee \;
\neg {\Box}_a P\big( C_{10}(a,b) \wedge C_{11}(a,c) \wedge C_{13}(c) \big) 
\label{eq:ac15}
\end{equation}
The first disjunct is clearly true, because $a$ can rationally believe that it is {\em possible} that hitting the brake is consistent with avoiding a rear-end collision and therefore with any planned action $C_{13}(c)\Rightarrow_c A_9(c)$, even if this is improbable.  So action plan (\ref{eq:ac10}) does not violate joint autonomy.  

We now check whether a failure to hit the brake, as in action plan (\ref{eq:ac11}), is inconsistent with the pedestrian's action plan $C_{12}(b)\Rightarrow_b A_8(b)$.  There is no violation of autonomy if
\begin{equation}
{\dia}_a P \big( \neg A_7(a) \wedge A_8(b) \big) \; \vee \;
\neg {\Box}_a P\Big( C_{10}(a,b) \wedge C_{11}(a) \wedge C_{12}(b) \Big) 
\label{eq:ac16}
\end{equation}
The first disjunct of (\ref{eq:ac16}) is clearly false for $b$'s action plan $C_{12}(b)\Rightarrow_b A_8(b)$, because driver $a$ cannot rationally believe that a failure to hit the brake is consistent with it.  The second disjunct is likewise false, because driver $a$ has no reason to believe that $C_{10}(a,b)$, $C_{11}(a,c)$ and $C_{12}(b)$ are mutually inconsistent.  Thus (\ref{eq:ac16}) is false, and we have a violation of autonomy.  The driver should therefore slam on the brake.  There is no need to check the other ethical principles, because only one of the possible action plans satisfies the autonomy principle.

\subsection{Implementation issues}

\red{While it is not our purpose to address engineering aspects of deontically-grounded VA, we can take note of some implementation issues that arise.  The main implication of our proposal is that the portion of an AI system that makes ethically relevant decisions must be rule-based (i.e., an instance of GOFAI) because it must consist of action plans.  Fortuitously, action plans have an if--then structure that is convenient for coding rules.}

\red{One can ask whether a rule-based system is adequate for the complexities of real-life decision making, but this is, of course, a problem that is not confined to deontically-based VA.  We do not attempt here to judge the versatility of rule-based AI, but we note that it seems to be increasingly viewed as technically viable and even necessary due to the nontransparency of deep learning and support vector machines.  Regarding autonomous vehicles, for example, \citeauthor{Bra18} (\citeyear{Bra18}) states, ``Many companies have shifted to rule-based AI, an older technique that lets engineers hard-code specific behaviors or logic into an otherwise self-directed system.''  The technical community has ample experience at accurately coding and debugging huge rule-based systems.   An ordinary (non-self-driving) automobile is already regulated by more than 100,000 lines of code.  Ethics-based systems can evolve through several versions and be updated as necessary, as with any other type of complex software.   Rule-based AI can also be combined with machine learning \citep{WazPol20}.  Even in a pure ML system, it is possible to derive rules that approximate the directives generated by ML \citep{SoaAngCos20} and perhaps subject them to ethical evaluation.}

\red{The test propositions used to evaluate the ethical status of action plans need not appear in the AI rule base, and it is a further implementation decision whether to generate them automatically.  This is fairly straightforward (less so for the utilitarian test), because the procedure for doing so can be clearly specified as shown above.  Machine learning and other forms of empirical VA can then be used to evaluate the truth of the test propositions.  }

\section{Conclusion}

As AI inexorably enters everyday life, it takes a seat alongside human persons.  AI's increasing sophistication bestows power, and power begets  responsibility.  Humanity's goal should be to invest machines with a moral sensitivity that mimics the human conscience. But conscience is dynamic rather than static, and adjusts ethical principles systematically to empirical observations.  In this paper we have elaborated two challenges to AI moral reasoning that spring from the interrelation of facts and values.  The first is a confusion that mistakenly identifies facts for values; the second is a confusion that misunderstands the process of moral reasoning.  In addressing these challenges, we have identified  instances of how and why AI can commit the naturalistic fallacy, of moving illicitly from ``is's'' to ``oughts,''  and doing so oversimplifies the process of moral reasoning.  We have sketched, in response, a proposal for understanding moral reasoning in machines, one that highlights how deontological ethical principles can interact with factual states of affairs.   




\end{document}